\documentclass{llncs}

\usepackage{graphicx}
\usepackage{amsmath,amssymb} 
\usepackage{color}
\usepackage[width=122mm,left=12mm,paperwidth=146mm,height=193mm,top=12mm,paperheight=217mm]{geometry}
\usepackage{graphicx}
\usepackage[font={small},skip=-0.5mm]{caption}
\usepackage[font={small},belowskip=-0.5mm]{subcaption}
\usepackage[dvipsnames]{xcolor}
\usepackage{tabularx}
\usepackage{xcolor}
\usepackage{booktabs}
\usepackage{adjustbox}
\usepackage{multirow}
\usepackage{wrapfig}
\usepackage{multicol}

\usepackage{tikz}
\usetikzlibrary{shapes.callouts}
\usetikzlibrary{arrows}
\usetikzlibrary{chains, fit, quotes, automata, arrows}
\usetikzlibrary{matrix,positioning}

\def\etal{\emph{et al. }}
\begin{document}

\title{Y-Net: Joint Segmentation and Classification for Diagnosis of Breast Biopsy Images}
\titlerunning{Hamiltonian Mechanics}  
%
\author{Sachin Mehta\inst{1} \and
Ezgi Mercan\inst{1} \and Jamen Bartlett\inst{2} \and Donald Weaver\inst{2} \and Joann G. Elmore\inst{3} \and Linda Shapiro\inst{1}}
\authorrunning{S. Mehta et al.} 
\institute{University of Washington, Seattle WA 98195, USA \\
 \and
University of Vermont, Burlington VT 05405, USA\\
\and
University of California, Los Angeles CA 90095, USA \\
\email{\{sacmehta, ezgi, shapiro\}@cs.washington.edu \{jamen.bartlett, donald.weaver\}@uvmhealth.org}  jelmore@mednet.ucla.edu
}

\maketitle              

\begin{abstract}
In this paper, we introduce a conceptually simple network for generating discriminative tissue-level segmentation masks for the purpose of breast cancer diagnosis. Our method efficiently segments different types of tissues in breast biopsy images while simultaneously predicting a discriminative map for identifying important areas in an image. Our network, Y-Net, extends and generalizes U-Net by adding a parallel branch for discriminative map generation and by supporting convolutional block modularity, which allows the user to adjust network efficiency without altering the network topology. Y-Net delivers state-of-the-art segmentation accuracy while learning $6.6\times$ fewer parameters than its closest competitors. The addition of descriptive power from Y-Net's discriminative segmentation masks improve diagnostic classification accuracy by 7\% over state-of-the-art methods for diagnostic classification. Source code is available at: \url{https://sacmehta.github.io/YNet}.
\end{abstract}

\section{Introduction}
Annually, millions of women depend on pathologists' interpretive accuracy to determine whether their breast
biopsies are benign or malignant \cite{Fine2006}. Diagnostic errors are alarmingly frequent, lead to incorrect treatment
recommendations, and can cause significant patient harm \cite{elmore2015diagnostic}.
Pathology as a field has been slow to
move into the digital age, but in April 2017, the FDA authorized the marketing of the Philips IntelliSite Pathology
Solution (PIPS), the first whole slide imaging system for interpreting digital surgical pathology slides on the basis
of biopsy tissue samples, thus changing the landscape\footnote{\url{ https://www.fda.gov/NewsEvents/Newsroom/PressAnnouncements/ucm552742.htm}}.

Convolutional neural networks (CNNs) produce state-of-the-art results in natural \cite{zhao2017pyramid,he2016deep} and biomedical classification and segmentation \cite{hou2016patch,ronneberger2015u} tasks. Training CNNs directly on whole slide images (WSIs) is difficult due to their massive size. Sliding-window-based approaches for classifying \cite{hou2016patch,geccer2016detection} and segmenting \cite{ronneberger2015u,mehta2017learning} medical images have shown promising results. Segmentation and classification are usually separate steps in automated diagnosis systems.

Segmentation-based methods consider tissue structure, such as size and distribution, to help inform class boundary decisions. However, these segmentation methods suffer from two major drawbacks. First, labeled data is scarce because the labeling of biopsy images is time-consuming and must be done by domain experts. Second, segmentation-based approaches are not able to weigh the importance of different tissue types. The latter limitation is particularly concerning in biopsy images, because not every tissue type in biopsy images is relevant for cancer detection. On the other hand, though classification-based methods fail to provide structure- and tissue-level information, they can identify regions of interest inside the images that should be used for further analysis.
\begin{figure}[b!]
\centering
\includegraphics[height=58px]{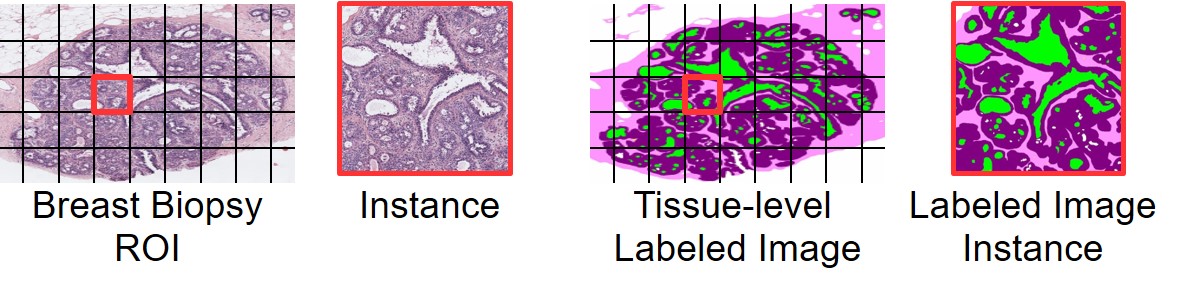}
\includegraphics[height=58px]{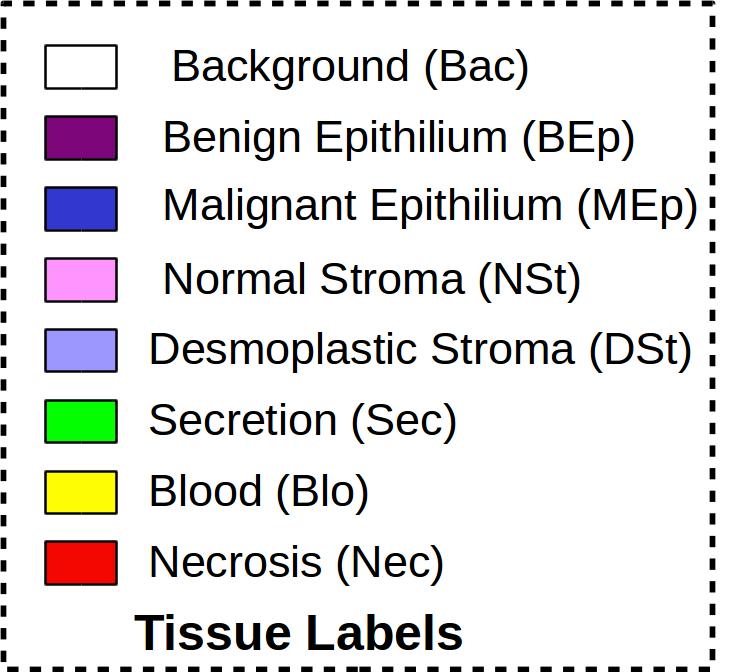}
\caption{This figure shows  (at left) the breast biopsy ROI with H\&E staining broken into multiple instances with one instance enlarged to show more detail. On the right are the pixel-wise tissue-level labelings of the ROI and the instance.}
\label{fig:wsiInstance}
\end{figure}

In this paper, we combine the two different methods, segmentation and classification, and introduce a new network called Y-Net that simultaneously generates a tissue-level segmentation mask and a discriminative (or saliency) map. Y-Net generalizes the U-Net network \cite{ronneberger2015u}, a well-known segmentation network for biomedical images. Y-net includes a \textit{plug-and-play} functionality that enables the use of different types of convolutional blocks without changing the network topology, allowing users to more easily explore the space of networks and choose more efficient networks. For example, Y-Net delivers the same segmentation performance as that of \cite{mehta2017learning} while learning $6.6 \times$ fewer parameters. Furthermore, the discriminative tissue-level segmentation masks produced using Y-Net provide powerful features for diagnosis. Our results suggest that Y-Net is 7\% more accurate than state-of-the-art segmentation and saliency-based methods \cite{mehta2017learning,geccer2016detection}.

\noindent \textbf{Statement of problem:} The problem we wish to solve is the simultaneous segmentation and diagnosis of whole slide breast cancer biopsy images. For this task, we used the breast biopsy dataset in \cite{elmore2015diagnostic,mehta2017learning} that consists of 240 whole slide breast biopsy images with heamatoxylin and eosin (H\&E) staining. A total of 87 pathologists diagnosed a randomly assigned subset of 60 slides into four diagnostic categories (benign, atypia, ductal carcinoma {\it in situ}, and invasive cancer), producing an average of 22 diagnostic labels per case. Then, each slide was carefully interpreted by a panel of three expert pathologists to assign a consensus diagnosis for each slide that we take to be the gold standard ground truth. Furthermore, the pathologists have marked 428 regions of interest (ROIs) on these slides that helped with diagnosis and a subset of 58 of these ROIs have been hand segmented by a pathology fellow into eight different tissue classifications: \textit{background, benign epithelium, malignant epithelium, normal stroma, desmoplastic stroma, secretion, blood,} and \textit{necrosis}. The average size of these ROIs is $10,000 \times 12,000$. We use these 428 ROIs for our data set. 

In this work, we break each ROI into a set (or bag) of equal size patches that we will call {\it instances}, as shown in Figure \ref{fig:wsiInstance}. Each ROI $X$ has a known groundtruth diagnostic label $Z$. There are no separate diagnostic labels for the instances; they all have the same groundtruth label $Z$, but some of them contribute to the diagnosis of $Z$ and others do not. Our system, therefore, will learn the {\it discriminativeness} of each instance during its analysis. Furthermore, each pixel of each of these instances has a known tissue classification into one of the eight categories; tissue classification must be learned from the groundtruth ROIs. Using the groundtruth diagnostic labels $Z$ of the ROIs and the groundtruth tissue labels $Y$ from the 58 labeled ROIs, our goal is to build a classification system that can input a ROI, perform simultaneous segmentation and classification, and output a diagnosis. Our system, once trained, can be easily applied to WSIs.
\begin{figure}[b!]
\centering
\includegraphics[width=\columnwidth]{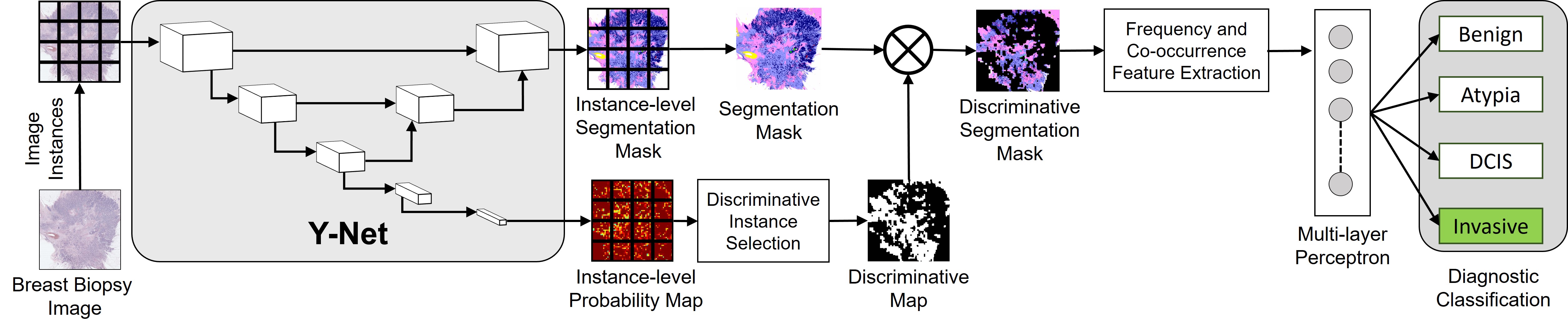}
\caption{Overview of our method for detecting breast cancer.}
\label{fig:overview}.
\end{figure}

\noindent \textbf{Related work:} Biomedical images are difficult to classify and segment, because their anatomical structures vary in shape and size. CNNs, by virtue of their representational power and capacity for capturing structural information, have made such classification and segmentation tasks possible \cite{ronneberger2015u,hou2016patch}. The segmentation-based method in \cite{mehta2017learning} and saliency map-based method in \cite{geccer2016detection} are most similar to our work. Mehta \etal \cite{mehta2017learning} developed a CNN-based method for segmenting breast biopsy images that produces a tissue-level segmentation mask for each  WSI. The histogram features they extracted from the segmentation masks were used for diagnostic classification. Ge{\c{c}}er \cite{geccer2016detection} proposed a saliency-based method for diagnosing cancer in breast biopsy images that identified relevant regions in breast biopsy WSIs to be used for diagnostic classification. Our main contribution in this paper is a method for \textit{joint learning} of both segmentation and classification. Our experiments show that joint learning improves diagnostic accuracy.

\section{A System for Joint Segmentation and Classification}
Our system (Figure \ref{fig:overview}) is given an ROI from a breast biopsy WSI and breaks it into instances that are fed into Y-Net. Y-Net produces two different outputs: an instance-level segmentation mask and an instance-level probability map. The instance-level segmentation masks have, for each instance, the predicted labels of the eight different tissue types. These are combined to produce a segmentation mask for the whole ROI. The instance-level probability map contains (for every pixel) the maximum value of the probability of that instance being in one of the four diagnostic categories. This map is thresholded to binary and combined with the segmentation mask to produce the discriminative segmentation mask. A multi-layer perceptron then uses the frequency and co-occurrence features extracted from the final mask to predict the cancer diagnosis.

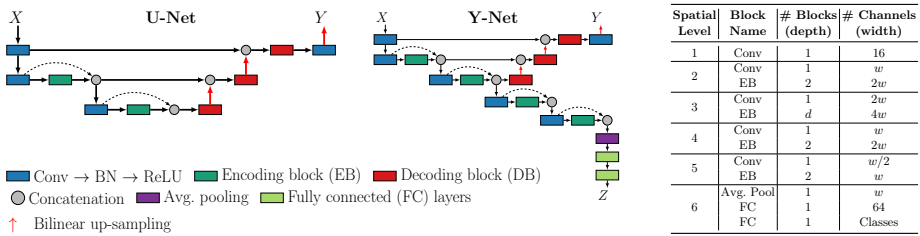
\begin{figure}[b!]
\centering
\begin{subfigure}[b]{0.67\columnwidth}
\resizebox{\columnwidth}{!}{
\hspace*{-70pt}
	\noindent
    \definecolor{c1}{RGB}{31,120,180}
\definecolor{c2}{RGB}{27,158,119}
\definecolor{c3}{RGB}{186,186,186}
\definecolor{c4}{RGB}{215,25,28}
\definecolor{c5}{RGB}{128,177,211}
\definecolor{c6}{RGB}{123,50,148}
\definecolor{c7}{RGB}{166,217,106}

\tikzset{>=latex}

\newcommand\rig{0.4cm}
\newcommand\lef{1cm}
\newcommand\belowRig{0.5cm}
\newcommand\lw{0.5mm}
\newcommand\lwA{0.9mm}
\newcommand\lwEncSh{0.5mm}
\newcommand\lwBil{0.5mm}
\newcommand\belowCl{1cm}

\newcommand\minH{0.5cm}

\newcommand{\stageTwo}{
  \begin{tikzpicture}[
block/.style={
text width={width("CONV")},
minimum height=\minH,
align=left,
line width=\lw,
font=\LARGE}]
	\node[block] (x) {{\Huge $X$}};
    
    \node[block,draw, fill=c1, below=1cm of x] (down1) {};
    \node[block,draw, fill=c1, below=1cm of down1] (down2) {};
    \node[block,draw, fill=c2, right=1cm of down2] (eb1) {};
    \node[circle,line width=\lw, draw, fill=c3, right=1cm of eb1, minimum height=\minH] (con1) {};

    \node[block,draw, fill=c1, below=1cm of con1] (down3) {};
    \node[block,draw, fill=c2, right=1cm of down3] (eb3) {};
    \node[circle,line width=\lw, draw, fill=c3, right=1cm of eb3, minimum height=\minH] (con2) {};
    \node[block,draw, fill=c4, right=1cm of con2] (sp1) {};
    
    \node[circle, line width=\lw, draw, fill=c3, above=1cm of sp1, minimum height=\minH] (con3) {};
    \node[block,draw, fill=c4, right=1cm of con3] (sp2) {};
    
    \node[circle, line width=\lw, draw, fill=c3, above=1cm of sp2, minimum height=\minH] (con4) {};
    \node[block,draw, fill=c4, right=1cm of con4] (sp3) {};
    \node[block,draw, fill=c1, right=1cm of sp3] (seg) {};
    
    \node[block, above=1cm of seg] (y) {{\Huge $Y$}};
    
    \path (x) -- (y) node[midway] (b) {\Huge \textbf{U-Net}};
    
    \draw[->, line width=\lwA] (x) -- (down1);
    \draw[->, line width=\lwA] (down1) -- (down2);
    \draw[->, line width=\lwA] (down2) -- (eb1);
    \draw[->, line width=\lwA] (eb1) -- (con1);
    
    \draw[->, line width=\lwA] (con1) -- (down3);
    \draw[->, line width=\lwA] (down3) -- (eb3);
    \draw[->, line width=\lwA] (eb3) -- (con2);
    \draw[dashed, ->, line width=\lwEncSh] (down2) to[out=40, in=145] (con1.north west);
    \draw[dashed, ->, line width=\lwEncSh] (down3) to[out=40, in=145] (con2.north west);
    \draw[->, line width=\lwA] (con2) -- (sp1);
    
    \draw[->,red, line width=1mm] (sp1) -- (con3);
    \draw[->, line width=\lwA] (con1) -- (con3);
    \draw[->, line width=\lwA] (con3) -- (sp2);
    
    \draw[->,red, line width=1mm] (sp2) -- (con4);
    \draw[->, line width=\lwA] (down1) -- (con4);
    \draw[->, line width=\lwA] (con4) -- (sp3);
    \draw[->, line width=\lwA] (sp3) -- (seg);
    
    \draw[->,red, line width=1mm] (seg) -- (y);
    
    \node[block,draw, fill=c1, below=6cm of down1] (desc0) {};
    \node[right=0.1cm of desc0] (desc10) {{\Huge\text{Conv $\rightarrow$ BN $\rightarrow$ ReLU}}};
    \node[block,draw, fill=c2, right=0.5cm of desc10] (desc2) {};
    \node[right=0.1cm of desc2] (desc22) {{\Huge Encoding block (EB)}};
    
    \node[block,draw, fill=c4, right=0.5cm of desc22] (desc3) {};
    \node[right=0.1cm of desc3] (desc30) {{\Huge Decoding block (DB)}};
    \node[circle,draw, fill=c3, below=0.5cm of desc0, minimum height=0.7cm] (desc4) {};
    \node[right=0.1cm of desc4] (desc44) {{\Huge Concatenation}};
    \node[block,draw, fill=c6, right=0.5cm of desc44] (desc5) {};
    \node[right=0.1cm of desc5] (desc55) {{\Huge Avg. pooling}};
    
    \node[block,draw, fill=c7, right=0.5cm of desc55] (desc00) {};
    \node[right=0.1cm of desc00] (ddd) {{\Huge Fully connected (FC) layers}};
    \node[block, below=0.5cm of desc4] (desc6) {{\Huge\textcolor{red}{\textbf{$\uparrow$}}}};
    \node[right=0.1cm of desc6] (desc66) {{\Huge Bilinear up-sampling}};

    \node[block, right=2cm of y] (x) {{\huge $X$}};
    \node[block,draw, fill=c1, below=\belowRig of x] (down1) {};
    \node[block,draw, fill=c1, below=\belowRig of down1] (down2) {};
    \node[block,draw, fill=c2, right=\rig of down2] (eb1) {};
    \node[circle,line width=\lw, draw, fill=c3, right=\rig of eb1, minimum height=\minH] (con1) {};
    \draw[dashed, ->, line width=\lwEncSh] (down2) to[out=40, in=145] (con1.north west);
    
    \node[block,draw, fill=c1, below=\belowRig of con1] (down3) {};
    \node[block,draw, fill=c2, right=\rig of down3] (eb3) {};
    \node[circle,line width=\lw, draw, fill=c3, right=\rig of eb3, minimum height=\minH] (con2) {};
    \draw[dashed, ->, line width=\lwEncSh] (down3) to[out=40, in=145] (con2.north west);
    \node[block,draw, fill=c4, right=\rig of con2] (sp1) {};
    
    \node[circle, line width=\lw, draw, fill=c3, above=\belowRig of sp1, minimum height=\minH] (con3) {};
    \node[block,draw, fill=c4, right=\rig of con3] (sp2) {};
    
    \node[circle, line width=\lw, draw, fill=c3, above=\belowRig of sp2, minimum height=\minH] (con4) {};
    \node[block,draw, fill=c4, right=\rig of con4] (sp3) {};
    \node[block,draw, fill=c1, right=\rig of sp3] (seg) {};
    
    \node[block, above=\belowRig of seg] (y) {{\huge $Y$}};
    
    \path (x) -- (y) node[midway] (b) {\Huge \textbf{Y-Net}};
    
    \draw[->, line width=\lw] (x) -- (down1);
    \draw[->, line width=\lw] (down1) -- (down2);
    \draw[->, line width=\lw] (down2) -- (eb1);
    \draw[->, line width=\lw] (eb1) -- (con1);
    
    \draw[->, line width=\lw] (con1) -- (down3);
    \draw[->, line width=\lw] (down3) -- (eb3);
    \draw[->, line width=\lw] (eb3) -- (con2);
    \draw[->, line width=\lw] (con2) -- (sp1);
    
    \draw[->,red, line width=\lwBil] (sp1) -- (con3);
    \draw[->, line width=\lw] (con1) -- (con3);
    \draw[->, line width=\lw] (con3) -- (sp2);
    
    \draw[->,red, line width=\lwBil] (sp2) -- (con4);
    \draw[->, line width=\lw] (down1) -- (con4);
    \draw[->, line width=\lw] (con4) -- (sp3);
    \draw[->, line width=\lw] (sp3) -- (seg);
    
    \draw[->,red, line width=\lwBil] (seg) -- (y);
    
    \node[block,draw, fill=c1, below=6mm of con2] (down4) {};
    \node[block,draw, fill=c2, right=\rig of down4] (eb8) {};
    \node[circle, line width=\lw, draw, fill=c3, right=\rig of eb8, minimum height=\minH] (con5) {};
    \draw[dashed, ->, line width=\lwEncSh] (down4) to[out=40, in=145] (con5.north west);
    
    \node[block,draw, fill=c1, below=\rig of con5] (down5) {};
    \node[block,draw, fill=c2, right=\rig of down5] (eb10) {};
    \node[circle, line width=\lw, draw, fill=c3, right=\rig of eb10, minimum height=\minH] (con6) {};
    \draw[dashed, ->, line width=\lwEncSh] (down5) to[out=40, in=145] (con6.north west);
    
    \node[block,draw, fill=c6, below=\rig of con6] (glb) {};
    \node[block,draw, fill=c7, below=\rig of glb] (fc1) {};
    \node[block,draw, fill=c7, below=\rig of fc1] (fc2) {};
    \node[block, below=\rig of fc2] (yc) {{\huge $Z$}};
    
    \draw[->, line width=\lw] (con2) -- (down4);
    \draw[->, line width=\lw] (down4) -- (eb8);
    \draw[->, line width=\lw] (eb8) -- (con5);
    
    \draw[->, line width=\lw] (con5) -- (down5);
    \draw[->, line width=\lw] (down5) -- (eb10);
    \draw[->, line width=\lw] (eb10) -- (con6);
    
    \draw[->, line width=\lw] (con6) -- (glb);
    \draw[->, line width=\lw] (glb) -- (fc1);
    \draw[->, line width=\lw] (fc1) -- (fc2);
    \draw[->, line width=\lw] (fc2) -- (yc);
    
 \end{tikzpicture}
}\stageTwo
}
\caption{U-Net vs. Y-Net}
\label{fig:unet}
\end{subfigure}
\hfill
\begin{subfigure}[b]{0.28\columnwidth}
\resizebox{\columnwidth}{!}{
\begin{tabular}{c|c|c|c}
\toprule
\textbf{Spatial} &  \textbf{Block} & \textbf{\# Blocks}  & \textbf{\# Channels}\\
\textbf{Level} &  \textbf{Name} & \textbf{(depth)}  & \textbf{(width)}\\
\midrule
1 & Conv & 1 &  $16$\\
\hline
\multirow{2}{*}{2} & Conv & 1 & $w$\\
 & EB & 2 &  $2w$\\
\hline
\multirow{2}{*}{3} & Conv & 1 &  $2w$\\
 & EB & $d$ &  $4w$\\
\hline
\multirow{2}{*}{4} & Conv & 1 & $w$\\
 & EB & $2$ &   $2w$\\
\hline
\multirow{2}{*}{5} & Conv & 1 &  $w/2$\\
 & EB & $2$ &  $w$\\
\hline
\multirow{3}{*}{6} & Avg. Pool & 1 &  $w$\\
 & FC & 1 & 64\\
 & FC & 1 &  Classes\\
\bottomrule
\end{tabular}
}
\caption{Encoding Network}
\label{fig:encodingNet}
\end{subfigure}
\caption{ (a) Comparison between U-Net and Y-Net architectures. (b) The encoding network architecture used in (a). U-Net in (a) is a generalized version of U-Net \cite{ronneberger2015u}.} 
\label{fig:netCOmpare}
\end{figure}

\subsection{Y-Net Architecture} 
\label{ssec:netArch}
Y-Net is conceptually simple and generalizes U-Net \cite{ronneberger2015u} to joint segmentation and classification tasks. U-Net outputs a single segmentation mask. Y-Net adds a second branch that outputs the classification label. The classification output is distinct from the segmentation output and requires feature extraction at low spatial resolutions. We first briefly review the U-Net architecture and then introduce the key elements in the Y-Net architecture.

\noindent \textbf{U-Net:} U-Net is composed of two networks: (1) encoding network and (2) decoding network. The encoding network can be viewed as a stack of encoding and down-sampling blocks. The encoding blocks learn input representations; down-sampling helps the network learn scale invariance. Spatial information is lost in both convolutional and down-sampling operations.  The decoder can be viewed as a stack of up-sampling and decoding blocks. The up-sampling blocks help in inverting the loss of spatial resolution, while the decoding blocks  help the network to compensate for the loss of spatial information in the encoder. U-Net introduces skip-connections between the encoder and the decoder, which enables the encoder and the decoder to share information.  

\noindent \textbf{Y-Net:} Y-Net (Figure \ref{fig:unet}) adopts a two-stage procedure. The first stage outputs the instance-level segmentation mask, as U-Net does, while the second stage adds a parallel branch that outputs the instance-level classification label. In spirit, our approach follows Mask-RCNN \cite{he2017mask} which jointly learns the segmentation and classification of natural images. Unlike Mask-RCNN, Y-Net is fully convolutional; that is, Y-Net does not have any region proposal network. Furthermore, training Y-Net is different from training Mask-RCNN, because Mask-RCNN is trained with object-level segmentations and classification labels. Our system has diagnostic labels for entire ROIs, but not for the instance-level.

Y-Net differs from the U-Net in the following aspects:

\noindent \textbf{\textit{Abstract representation of encoding and decoding blocks:}} At each spatial level, U-Net uses the same convolutional block (a stack of convolutional layers) in both the encoder and the decoder. Instead, Y-Net abstracts this representation and represents convolutional blocks as general encoding and decoding blocks that can be used anywhere, and are thus not forced to be the same at each spatial level. Representing Y-Net in such a modular form provides it a \textit{plug-and-play} functionality and therefore enables the user to try different convolutional block designs without changing the network topology.

\noindent \textbf{\textit{Width and depth multipliers:}} Larger CNN architectures tend to perform better than smaller architectures. We introduce two hyper-parameters, a width multiplier $w$ and a depth-multiplier $d$, that allow us to vary the size of the network. These parameters allow Y-Net to span the network space from smaller to larger networks, allowing identification of better network structures. 

\noindent \textbf{\textit{Sharing features:}} While U-Net has skip-connections between the encoding and decoding stages, Y-net adds a skip-connection between the first and last encoding block at the same spatial resolution in the encoder, as shown in Figure \ref{fig:netCOmpare} with a dashed arrow, to help improve segmentation. 

\noindent \textbf{\textit{Implementation details:}}  The encoding network in Y-Net (Figure \ref{fig:unet}) consists of the repeated application of the encoding blocks and $3\times 3$ convolutional layers with a stride of 2 for down-sampling operations, except for the first layer which is a $7\times 7$ standard convolution layer with a stride of 2. Similarly, the decoding network in Y-Net consists of the repeated application of the decoding blocks and bilinear up-sampling for up-sampling operations. We first train Y-Net  for segmentation and then attach the remaining encoding network (spatial levels 4, 5 and 6 in Figure \ref{fig:encodingNet}) to jointly train for segmentation and classification. We define a multi-task loss on each instance as $\mathcal{L} = \mathcal{L}_{seg} + \mathcal{L}_{cls}$, where $\mathcal{L}_{seg}$ and $\mathcal{L}_{cls}$ are the multi-nominal cross-entropy loss functions for the segmentation and classification tasks, respectively.  All layers and blocks, except the classification and fully connected (FC) layers, are followed by a batch normalization and ReLU non-linearity. An average pooling layer with adaptive kernel size enables Y-Net to deal with arbitrary image sizes. 

\subsection{Discriminative Instance Selection}
\label{ssec:discInSel}
The encoding network in Y-Net generates a $C$-dimensional output vector $\mathbf{z}$ of real values; $C$ represents the number of diagnostic classes. The real-values in $\mathbf{z}$ are normalized using a softmax function $\mathbf{\sigma}$ to generate another $C$-dimensional vector $\mathbf{\bar{z}} = \mathbf{\sigma(\mathbf{z})}$. It is reasonable to assume that instances with low probability will have low discriminativeness. If $\text{max}(\mathbf{\bar{z}}) > \tau$, then  the instance is considered {\it discriminative}, where $\tau$ is the threshold selected using the method in \cite{hou2016patch}. 

\subsection{Diagnostic Classification}
\label{ssec:diagClassIm}
Segmentation masks provide tissue-level information. Since training data with tissue-level information is limited and not all tissue types  contribute equally to diagnostic decisions, our system combines the segmentation mask with the discriminative map to obtain a  {\it tissue-level discriminative segmentation mask}. Frequency and co-occurrence histograms are extracted from the discriminative segmentation mask and used to train a multi-layer perceptron (MLP) with 256, 128, 64, and 32 hidden nodes to predict the diagnostic class. 

\section{Experiments}
In this section, we first study the effect of the modular design in Y-Net. We then compare the performance of Y-Net with state-of-the-art methods on tissue-level segmentation as well as on diagnostic classification tasks. For evaluation, we used the breast biopsy dataset \cite{elmore2015diagnostic,mehta2017learning} that consists of 428 ROIs with classification labels and 58 ROIs with tissue-level labels.

\subsection{Segmentation Results}
We used residual convolutional blocks (RCB) \cite{he2016deep} and efficient spatial pyramid blocks (ESP) \cite{mehtaAnnony} for encoding and decoding. Based on the success of PSPNet for segmentation \cite{zhao2017pyramid}, we added pyramid spatial pooling (PSP) blocks for decoding.

\noindent \textbf{Training details:} We split the 58 ROIs into nearly equal training (\# 30) and test (\# 28) sets. For training, we extracted $384 \times 384$ instances with an overlap of 56 pixels at different image resolutions. We used standard augmentation strategies, such as random flipping, cropping, and resizing, during training. We used a 90:10 ratio for splitting training data into training and validation sets. We trained the network for 100 epochs using SGD with an initial learning rate of 0.0001, decaying the rate by a factor of 2 after 30 epochs. We measured the accuracy of each model using mean Region Intersection over Union (mIOU).

\noindent \textbf{Segmentation studies}:  Segmentation results are given in Table \ref{tab:segResults} and Table \ref{tab:segCompareSOTA}. We make the following observations:

\begin{table}[b!]
\centering
\begin{subtable}[b]{0.63\columnwidth}
\resizebox{\columnwidth}{!}{
\begin{tabular}{c|cc|ccc|cc||cc||cc}
\toprule
\multirow{2}{*}{\textbf{Row}} & \multicolumn{2}{c}{\textbf{Encoding}} & \multicolumn{3}{|c|}{\textbf{Decoding}} & \multicolumn{2}{|c||}{\textbf{Feature}} & \textbf{\# Params} & \textbf{mIOU} & \textbf{\# Params} & \textbf{mIOU}\\
 & \multicolumn{2}{c}{\textbf{Block}} & \multicolumn{3}{|c|}{\textbf{Block}} & \multicolumn{2}{|c||}{\textbf{Sharing}} & (in million) & & (in million) & \\
\cline{2-12}
\# & \textbf{ESP} & \textbf{RCB} & \textbf{ESP} & \textbf{RCB} & \textbf{PSP} & \textbf{Add} & \textbf{Concat} & \multicolumn{2}{|c||}{$w=64$} & \multicolumn{2}{|c}{$w=128$}\\
\midrule
R1 & \checkmark &  & \checkmark &  &  &  &  & \textbf{0.49} & 30.39 & \textbf{1.95} & 35.23\\
R2 & \checkmark &  & \checkmark &  &  & \checkmark  &  & \textbf{0.49} & 32.12 & \textbf{1.95}  & 36.19\\
R3 & \checkmark &  & \checkmark &  &  &  & \checkmark  & 0.57 & 34.58 & 2.25 & 38.03\\
\midrule
R4 &  &  \checkmark&  & \checkmark &  &  &  & 1.72 & 33.05 & 6.84 & 37.93\\
R5 & & \checkmark &  & \checkmark &  & \checkmark  &  & 1.72 & 36.34 & 6.84 & 39.21\\
R6 &  &  \checkmark & & \checkmark &  &  & \checkmark  & 1.81 & 38.75 & 7.16 & 40.23\\
\midrule
R7 & \checkmark &  &  &  & \checkmark  &   & \checkmark &  0.69 &  37.12 & 2.75 & 44.03 \\
R8 &  & \checkmark &  &  & \checkmark &  & \checkmark  &  1.91 & \textbf{41.96} & 7.62 & \textbf{44.19}\\
 \bottomrule
\end{tabular}
}
\caption{Network width vs. accuracy}
\label{tab:accAbla}
\end{subtable}
\hfill
\begin{subtable}[b]{0.36\columnwidth}
\centering
\includegraphics[width=0.9\columnwidth]{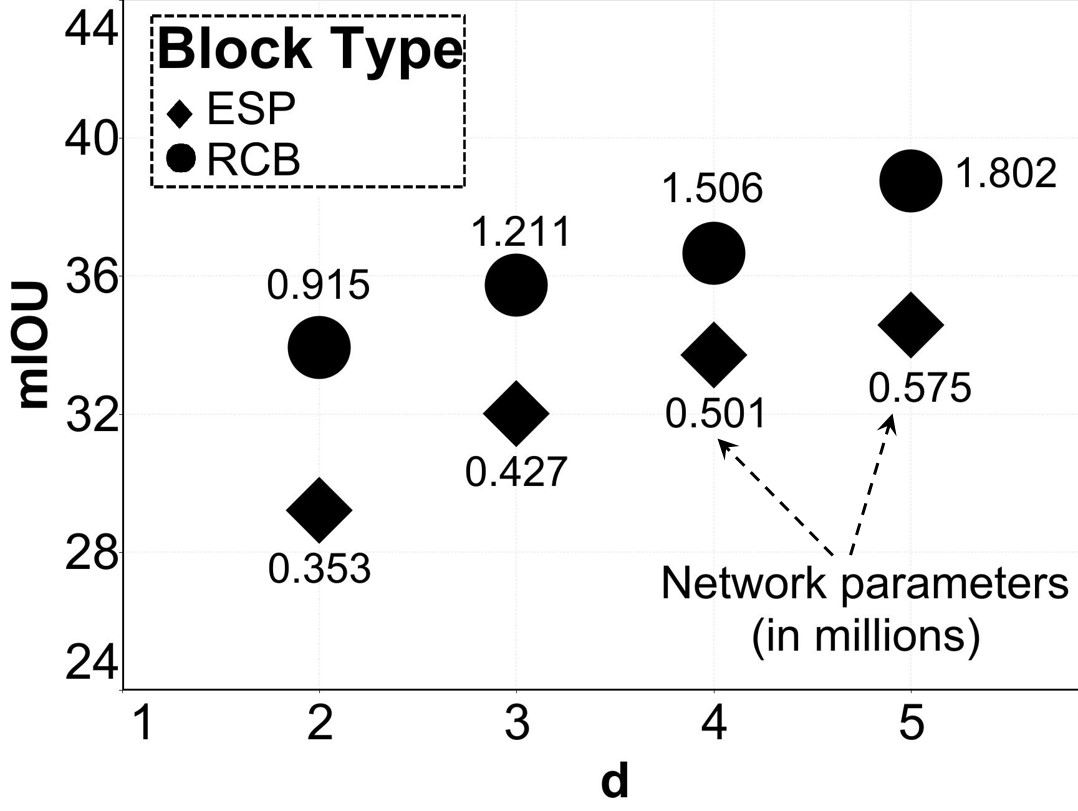}
\caption{Network depth vs. accuracy}
\label{tab:depth}
\end{subtable}
\caption{Ablation studies on Y-Net. In (a), we used $d=5$. In (b), we used $w=64$. The experimental settings in (b) were the same as for R3 and R6 in (a).}
\label{tab:segResults}
\end{table}

\noindent \textbf{\textit{Feature sharing:}} (Table \ref{tab:accAbla}) When features were shared between the encoding blocks at the same spatial level, the accuracy of the network improved by about 2\% (R2 and R5) with element-wise addition operations and about 4\% (R3 and R6) with concatenation operations for both ESP and RCB blocks.  The increase in number of parameters due to concatenation operations was not significant.

\noindent \textbf{\textit{Network depth}}: (Table \ref{tab:depth}) The value of $d$ was varied from 2 to 5 in Y-Net for both ESP and RCB types of convolutional blocks. The accuracy of the network increased with the depth of the network. When we increased $d$ from 2 to 5, the accuracy improved by about 4\% while the network parameters were increased by about $1.6\times$ and $1.9 \times$ for Y-Net with ESP and RCB respectively. In the following experiments, we used $d=5$. 

\noindent \textbf{\textit{Network width}}: (Table \ref{tab:accAbla}) When  the value of $w$ changed from 64 to 128, the accuracy of Y-Net with ESP (R1-R3) and RCB (R4-R6) increased by about 4\%. However, the number of network parameters increased drastically.

\noindent \textbf{\textit{PSP as decoding block}}: (Table \ref{tab:accAbla}) Changing the decoding block from ESP and RCB to PSP helped improve the accuracy by about 3\%. This is because the pooling operations in PSP modules helped the network learn better global contextual information. Surprisingly, when  the value of $w$ increased from 64 to 128, Y-Net with ESP and PSP delivered accuracies similar to PCB and PSP. This is likely due to the increased number of kernels per branch in the ESP blocks, which helps to learn better representations. Y-Net with ESP blocks learns about $3\times$ fewer parameters and is therefore more efficient.

\noindent \textbf{\textit{Joint Training}}: (Table \ref{tab:segCompareSOTA}) Training Y-Net jointly for both classification and segmentation tasks dropped the segmentation accuracy by about 1\%. This is likely because we trained the network using an instance-based approach and we did not have classification labels at instance-level.

\noindent \textbf{\textit{Comparison with state-of-the-art}}: (Table \ref{tab:segCompareSOTA}) Y-Net outperformed the plain \cite{badrinarayanan2017segnet} and residual \cite{fakhry2017residual} encoder-decoder networks by 7\% and 6\% respectively. With the same encoding block (RCB) as in \cite{mehta2017learning}, Y-Net delivered a similar accuracy while learning $2.85 \times$ fewer parameters. We note that Y-Net with ESP and PSP blocks also delivered a similar performance while learning $6.6 \times$ fewer parameters than \cite{mehta2017learning} and $2.77 \times$ fewer parameters than Y-Net with RCB and PSP blocks. Therefore, the modular architecture of Y-Net allowed us to explore different convolutional blocks with a minimal change in the network topology to find a preferred network design.

\begin{table}[t!]
\begin{subtable}[b]{0.45\columnwidth}
\centering
\resizebox{\columnwidth}{!}{
\begin{tabular}{l|c|c}
\toprule
\textbf{Network} & \textbf{mIOU} & \textbf{\# Params} (in million) \\
\midrule
Superpixel + SVM \cite{mehta2017learning} & 25.8 & NA \\
\hline
Badrinarayanan \etal \cite{badrinarayanan2017segnet} & 37.6 & 12.8 \\
Fakhry \etal \cite{fakhry2017residual} & 38.1 & 12.8\\
Mehta \etal \cite{mehta2017learning} & \textbf{44.20} & 26.03 \\
\midrule
YNet (ESP-PSP) - seg & 44.03 &  \textbf{2.75}\\
YNet (RCB-PSP) - seg & 44.19 &  7.62\\
\midrule
YNet (ESP-PSP) - joint & 43.24 &  3.91\\
YNet (RCB-PSP) - joint & 43.11 &  9.11\\
\bottomrule
\end{tabular}
}
\caption{Segmentation results}
\label{tab:segCompareSOTA}
\end{subtable}
\hfill
\begin{subtable}[b]{0.45\columnwidth}
\centering
\resizebox{0.7\columnwidth}{!}{
\begin{tabular}{lc}
\toprule
\textbf{Feature Type} & \textbf{Accuracy (in \%)} \\
\midrule
Pathologists (\# 44) & 70.0 \\
\midrule
LAB + LBP features \cite{geccer2016detection} & 45.0 \\
Segmentation mask \cite{mehta2017learning} & 54.5 \\
Saliency map \cite{geccer2016detection} & 55.0 \\
\midrule
\multicolumn{2}{c}{\textbf{Y-Net with different choices}}\\
Segmentation mask & 53.25 \\
\quad \quad -background & 52.22 \\
\quad \quad -stroma & 48.06 \\
Discriminative mask & \textbf{62.50} \\
\bottomrule
\end{tabular}
}
\caption{Diagnostic classification results}
\label{tab:diagAcc}
\end{subtable}
\caption{Comparison with state-of-the-art methods.  seg: training Y-Net only for the segmentation task; joint: joint learning for segmentation and classification tasks.}
\end{table}

\subsection{Diagnostic Classification Results}
\label{ssec:diagClass}
For classification experiments, we split the 428 ROIs in the dataset into almost equal training (\# 209) and test (\# 219) sets while maintaining the same class distribution across both the sets. We note that the 30 ROIs used for training the segmentation part were part of the training subset during the classification task. The tissue-level segmentation mask and discriminative map were first generated using Y-Net with ESP as encoding blocks and PSP as decoding blocks, which were then used to generate the discriminative segmentation mask. A 44-dimensional feature vector (frequency and co-occurrence histogram) was then extracted from the discriminative mask. These features were used to train a MLP that classifies the ROI into  four diagnoses (benign, atypia, DCIS, and invasive cancer). 

A summary of results is given in Table \ref{tab:diagAcc}. The classification accuracy improved by about 9\% when we used discriminative masks instead of segmentation masks. Our method outperformed state-of-the-art methods that use either the segmentation features \cite{mehta2017learning} or the saliency map \cite{geccer2016detection} by a large margin. Our method's 62.5\% accuracy is getting closer to the 70\% accuracy of 
trained pathologists in a study \cite{elmore2015diagnostic}. This suggests that the discriminative segmentation masks generated using Y-Net are powerful.

\section{Conclusion}
The Y-Net architecture achieved good segmentation and diagnostic classification accuracy on a breast biopsy dataset. Y-Net was able to achieve the same segmentation accuracy as state-of-the-art methods while learning fewer parameters. The features generated using discriminative segmentation masks were shown to be powerful and our method was able to attain higher accuracy than state-of-the-art methods. Though we studied breast biopsy images in this paper, we believe that Y-Net can be extended to other medical imaging tasks. 

\vspace{3mm}
\noindent \textbf{Acknowledgements:} Research reported in this publication was supported by the National Cancer Institute awards R01 CA172343, R01 CA140560, and RO1 CA200690. We would also like to thank NVIDIA Corporation for donating the Titan X Pascal GPU used for this research.

\bibliographystyle{splncs03}
\bibliography{main}

\end{document}